\documentclass[runningheads]{llncs}

\usepackage[mobile]{eccv}

\usepackage{eccvabbrv}

\usepackage{graphicx}
\usepackage{booktabs}
\usepackage[table]{xcolor}
\usepackage{colortbl}
\usepackage{array}
\usepackage{multirow}
\usepackage{amssymb}
\usepackage{pifont}
\usepackage{subcaption}
\usepackage{overpic}
\usepackage{tabularx}
\usepackage{float}

\definecolor{BarBlue}{RGB}{230,242,252}       
\definecolor{HeadPercep}{RGB}{225,242,233}    
\definecolor{HeadAlign}{RGB}{244,238,217}     
\definecolor{HeadReason}{RGB}{232,236,248}    
\definecolor{HeadVLN}{RGB}{251,232,232}       
\definecolor{CitySim}{RGB}{168,218,220}
\definecolor{WildSim}{RGB}{244,162,97}
\definecolor{lightgreen}{RGB}{220,240,220}
\usepackage[accsupp]{axessibility}  

\usepackage{hyperref}

\usepackage{orcidlink}

\makeatletter 
\newcommand{\printfnsymbol}[1]{%
  \textsuperscript{\@fnsymbol{#1}}%
} 

\makeatother

\begin{document}

\title{AirGroundBench: Probing Spatial Intelligence in Multimodal Large Models under Heterogeneous Multi-View Embodied Collaboration} 

\titlerunning{AirGroundBench}


\author{ Haotian Li\inst{1} \and Yida Wang\inst{1} \and Leyuan Wang\inst{1} \and Jinshan Lai\inst{4} \and Keyang Wang\inst{4} \and Zonghao Guo\inst{3} \and Qiang Ma\inst{3} \and Liuyu Xiang\inst{1} \and Jianwei Hu\inst{2}%
\thanks{Jianwei Hu and Zhaofeng He are co-corresponding authors.} \and Zhaofeng He\inst{1}\printfnsymbol{1} }

\authorrunning{H. Li et al.}

\institute{Beijing University of Posts and Telecommunications \and QiYuanLab \and Tsinghua University \and University of Electronic Science and Technology of China }

\maketitle

\begin{abstract}
  In recent years, multimodal large language models (MLLMs) have shown strong potential for embodied intelligence, yet their ability to maintain geometrically consistent spatial understanding across heterogeneous views remains under-evaluated. Existing benchmarks largely focus on single-agent, single-view perception, leaving a gap in systematic assessment for collaborative air-ground settings, where multi-scale observations are complementary but introduce scale mismatch, asymmetric occlusion, and reference-frame inconsistencies.We present \textbf{AirGroundBench}, a diagnostic benchmark for multi-view spatial intelligence in heterogeneous UAV-UGV collaboration. AirGroundBench is built from \textbf{11} high-fidelity simulated environments with \textbf{1,021} synchronized air-ground observation pairs, yielding \textbf{62k} dual-view \textbf{4-way single-choice} VQA instances and \textbf{115} closed-loop vision-language navigation (VLN) episodes. It covers \textbf{10} task types organized into four progressively demanding capability dimensions: spatial perception, cross-view alignment, spatial transformation and reasoning, and embodied decision-making. To support geometry-grounded evaluation and analysis, we provide structured spatial annotations, including cross-view object identities and metric 2D/3D bounding boxes.Evaluations on \textbf{13} representative MLLMs under UAV-only, UGV-only, and dual-view inputs reveal consistent bottlenecks: models perform relatively well on perception but struggle on alignment and transformation-heavy reasoning, and these deficits propagate to sequential decision-making in VLN. While dual-view inputs offer measurable gains over single-view variants, a persistent gap to human performance remains, highlighting geometric consistency as a key limitation for embodied MLLMs.
\end{abstract}

\begin{figure*}[t]
\centering

\newcommand{\figH}{4cm} 

\begin{minipage}{0.48\textwidth}
\centering
\includegraphics[height=\figH,keepaspectratio]{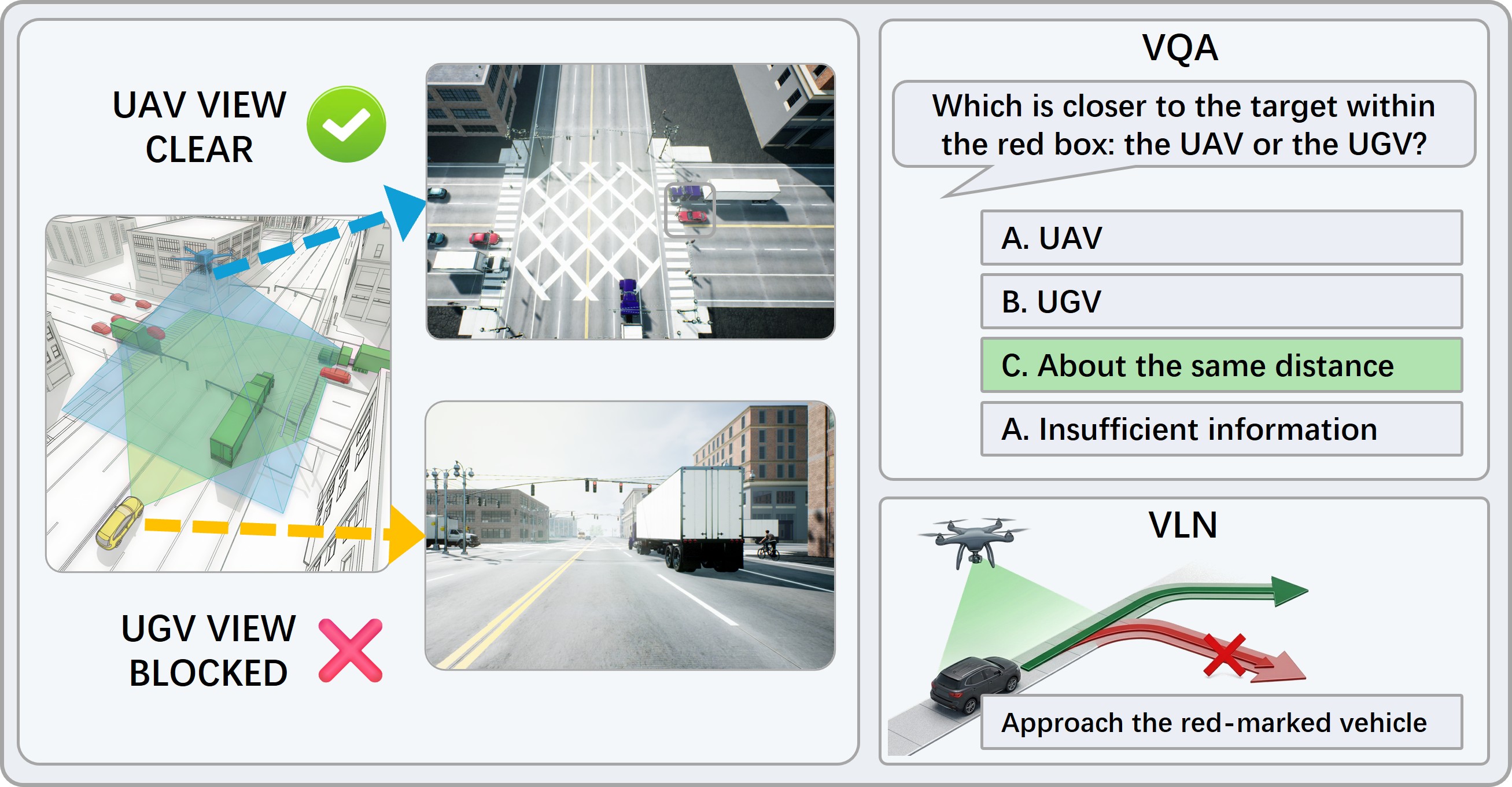}

\vspace{3pt}
{\small \hspace{20mm}(a)}
\end{minipage}
\hfill
\begin{minipage}{0.48\textwidth}
\raggedleft
\includegraphics[height=\figH,keepaspectratio]{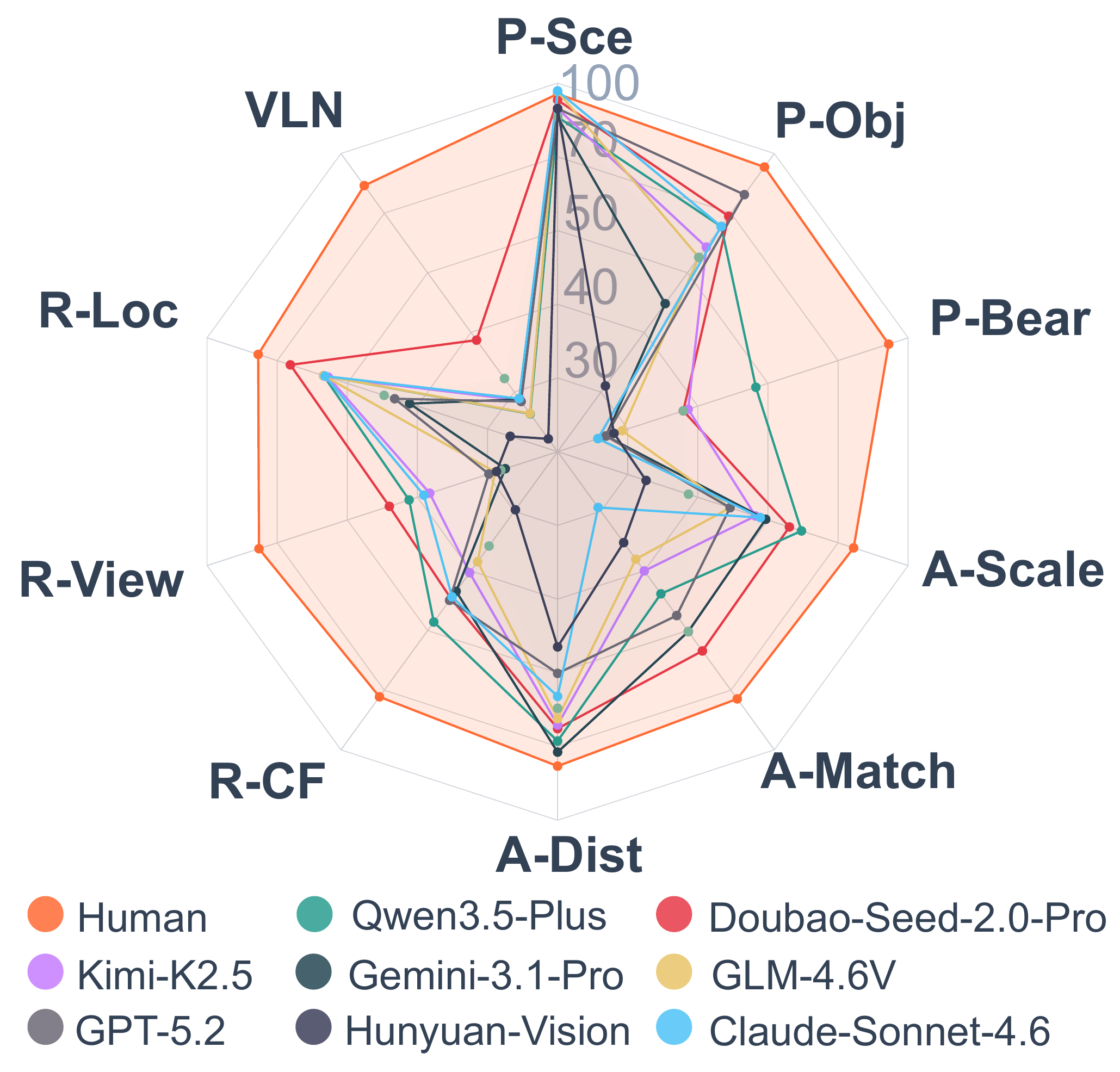}

\vspace{3pt}
\centerline{\small \hspace{18mm}(b)}
\end{minipage}

\caption{
\textbf{AirGroundBench: evaluating multi-view spatial intelligence for heterogeneous air--ground agents.}
(a) Motivation: heterogeneous UAV--UGV views often contain complementary but incomplete spatial observations, making cross-view reasoning necessary for accurate spatial interpretation and embodied task execution.
(b) Performance of representative MLLMs and humans on AirGroundBench across spatial capability dimensions, highlighting substantial gaps particularly in alignment and reasoning.
}

\label{fig:airgroundbench_intro}
\end{figure*}

\section{Introduction}

Recent Multimodal Large Language Models (MLLMs) have shown impressive capability in visual understanding and language-conditioned reasoning, motivating their use as general-purpose embodied agents. In embodied settings, however, success hinges not only on recognizing what is visible, but on maintaining geometric consistency---aligning observations across viewpoints and scales, preserving object identity, and reasoning over spatial relations under partial observability. Existing evaluations largely emphasize single-view spatial VQA or homogeneous embodied settings, leaving heterogeneous multi-view collaboration underexplored.

Heterogeneous air--ground systems provide a natural stress test. Aerial views supply global layout and long-range context, while ground views expose egocentric geometry and fine-grained semantics; critically, the two are complementary but not interchangeable. Their asymmetry introduces unavoidable challenges: scale mismatch, occlusion asymmetry, and reference-frame inconsistency. As illustrated in Fig.~\ref{fig:airgroundbench_intro}(a), accurate interpretation often requires binding the same entities across views and reconciling metric relations. Yet it remains unclear whether current MLLMs can reliably perform such cross-view alignment and transform it into robust decision-making.

To close this gap, we introduce \textbf{AirGroundBench}, a benchmark designed around geometric-consistency-centered multi-view spatial intelligence for heterogeneous air--ground agents. AirGroundBench is built from \textbf{11} evaluated simulated environments spanning diverse urban and wild structures, with \textbf{1,021} synchronized UAV--UGV image pairs. It contains \textbf{62k} dual-view VQA questions (each as a \textbf{four-option single-choice} problem) across nine task types, and \textbf{115} closed-loop vision--language navigation (VLN) episodes for sequential decision evaluation. Beyond RGB, we release depth/segmentation and simulator states to support geometry-verifiable annotation and future extensions.

Compared with prior spatial/embodied benchmarks (Table~\ref{tab:benchmark_comparison}), AirGroundBench uniquely combines (i) heterogeneous air--ground dual-view observations, (ii) explicit cross-view object identity and metric 3D grounding, and (iii) both static VQA and interactive VLN evaluation within one benchmark. This makes it possible to diagnose where geometric inconsistency arises (e.g., scale alignment vs.\ identity binding) and how it propagates to closed-loop failure.

Extensive evaluation over \textbf{13} representative MLLMs reveals a consistent capability ladder: performance drops sharply from perception to alignment and further to transformation-heavy reasoning, with the largest gaps in cross-view alignment and joint reasoning (Fig.~\ref{fig:airgroundbench_intro}(b)). In VLN, the gap becomes more consequential---models remain far below human success, indicating that small cross-view inconsistencies can accumulate into irreversible navigation errors.

Our contributions can be summarized as follows:
\begin{itemize}
    \item \textbf{Benchmark and taxonomy.} We propose a geometry-consistency-centered evaluation for heterogeneous air--ground multi-view spatial intelligence, covering four capability dimensions from perception and alignment to reasoning and embodied decision-making.
    \item \textbf{Dataset with verifiable geometry.} We release 11 evaluated environments with 1,021 synchronized UAV--UGV image pairs, 62k single-choice VQA questions, and 115 VLN episodes, together with cross-view object identities and metric 2D/3D annotations.
    \item \textbf{Large-scale evaluation and diagnosis.} We benchmark 13 MLLMs under dual-view and single-view ablations, and provide diagnostic analyses highlighting systematic failure modes in cross-view alignment and transformation.
\end{itemize}

\begin{table*}[t]
\centering
\caption{
\textbf{Comparison with existing spatial and embodied benchmarks.}
}
\resizebox{\textwidth}{!}{
\begin{tabular}{lcccccc}
\toprule
\textbf{Benchmark} & \textbf{Modality} & \textbf{Scale} & \textbf{VQA} & \textbf{VLN} & \textbf{View Setting} & \textbf{Metric Geometry} \\
\midrule
AirNav \cite{cai_airnav_2026} & RGB & 143k traj. & \ding{55} & \ding{51} & Single-view & \ding{55} \\
IndoorUAV \cite{liu_indooruav_2025} & RGB & 16k traj. & \ding{55} & \ding{51} & Single-view & \ding{55} \\
Embodied4C \cite{sohn_embodied4c_2025} & RGB & 1,149 VQA, 58 VLN tasks & \ding{51} & \ding{51} & Multi-view & \ding{55} \\
CityCube \cite{xu_citycube_2026} & RGB & 5,022 VQA & \ding{51} & \ding{55} & Multi-view & \ding{55} \\
SpatialMosaic \cite{lee_spatialmosaic_2025} & RGB & 1M VQA & \ding{51} & \ding{55} & Multi-view & \ding{51} \\
MM-UAVBench \cite{dai_mmuavbench_2025} & RGB, Video & 5,702 VQA & \ding{51} & \ding{55} & Multi-view & \ding{55} \\
AMVICC \cite{basappa_amvicc_2026} & RGB & 300 VQA & \ding{51} & \ding{55} & Dual-view & \ding{55} \\
AirCopBench \cite{zha_aircopbench_2025} & RGB, Point Cloud & 14.6k VQA & \ding{51} & \ding{55} & Multi-view & \ding{51} \\
\midrule
\noalign{\vskip-3pt}
\rowcolor{lightgreen!70}
\textbf{AirGroundBench (Ours)} & \textbf{RGB, Depth, Seg.} & \textbf{62k VQA, 115 traj.} & \ding{51} & \ding{51} & \textbf{Dual-view} & \ding{51} \\
\noalign{\vskip-1.6pt}
\bottomrule
\end{tabular}
}
\label{tab:benchmark_comparison}
\end{table*}

\section{Related Work}

\subsection{Spatial Understanding Evaluation for Multimodal Models}
Evaluating spatial understanding in multimodal large language models (MLLMs) has become an active research topic. Most existing evaluations adopt a VQA-style protocol: given an image (or a small set of images) and a question, models answer queries about spatial relations, object attributes, or scene layout. This format provides controlled diagnostics and aligns naturally with the perception--alignment--reasoning decomposition used in many recent spatial taxonomies.
Recent benchmarks broaden the spatial skill spectrum and provide more fine-grained evaluation. OmniSpatial~\cite{jia_omnispatial_2025} and MMSI-Bench~\cite{yang_mmsi-bench_2025} assess diverse spatial reasoning abilities, while STI-Bench~\cite{li_sti-bench_2025} further emphasizes spatial-temporal aspects. Multi-view evaluation is also receiving attention: ViewSpatial-Bench~\cite{li_viewspatial-bench_2025} studies cross-view spatial reasoning, and CityCube~\cite{xu_citycube_2026} targets large-scale urban spatial understanding.
Despite this progress, most benchmarks assume single-platform observations or homogeneous multi-camera rigs, where viewpoint distributions and metric scales are broadly consistent. Consequently, they do not explicitly probe heterogeneous viewpoint integration---for example, reconciling aerial and ground observations that exhibit systematic scale gaps, occlusion asymmetry, and reference-frame mismatch. Such heterogeneity makes geometric consistency a first-order challenge: models must align cross-view scales, associate identical entities across views, and preserve consistent metric relations (distance, direction) under severe viewpoint changes. Moreover, many existing datasets provide limited geometry-verifiable supervision, making it difficult to disentangle semantic guessing from genuine geometric alignment.

\subsection{Embodied AI Benchmarks}
Beyond static reasoning, embodied AI benchmarks evaluate spatial understanding under interaction and sequential decision-making. In robotics, cooperative perception studies how multiple agents fuse complementary observations under communication constraints, e.g., Where2comm~\cite{hu2022where2comm} and V2X-ViT~\cite{xu2022v2xvit}. These lines highlight that multi-agent observations can be complementary in coverage and occlusion, but their primary objective is perception fusion and system efficiency (e.g., detection accuracy under bandwidth limits) rather than diagnosing the spatial reasoning behaviors of general-purpose MLLMs or attributing failures to specific cognitive factors such as alignment vs. reasoning.
In the MLLM era, several embodied benchmarks have been proposed, including EmbodiedCity~\cite{gao_embodiedcity_2024} and Embodied4C~\cite{sohn_embodied4c_2025}. Collaborative settings such as CoPeD~\cite{zhou_coped-advancing_2024} and AirCopBench~\cite{zha_aircopbench_2025} further expand evaluation toward multi-agent scenarios. While valuable, existing embodied benchmarks rarely isolate cross-view geometric consistency as a first-class evaluation axis, and heterogeneous air--ground observation pairs are not systematically leveraged to stress-test scale alignment, entity binding across views, or viewpoint transformation under partial observability.

\subsection{Vision--Language Navigation Benchmarks}
Vision--Language Navigation (VLN) provides a canonical closed-loop testbed for grounded language understanding and planning. R2R~\cite{anderson2018r2r}, REVERIE~\cite{qi2020reverie}, and RxR~\cite{ku2020rxr} established widely used VLN formulations and datasets. Subsequent work improved long-horizon planning and spatial memory, such as DUET~\cite{chen2022duet} and GridMM~\cite{wang2023gridmm}, demonstrating the importance of structured spatial representations for instruction following.
With the rise of foundation models, NavGPT-2~\cite{zhou2024navgpt2} explores explicit reasoning with large vision--language models for navigation, and VLN-MME~\cite{zhao_vlnmme_2025} provides an evaluation suite for MLLMs in VLN. Aerial navigation benchmarks also emerge in parallel (e.g., AirNav~\cite{cai_airnav_2026}), reflecting viewpoint-specific challenges.
Nevertheless, most VLN benchmarks assume a single egocentric viewpoint at each step, and multi-view inputs are typically homogeneous (e.g., multiple cameras on one platform). This leaves open how heterogeneous multi-view observations can support navigation decisions (e.g., aerial global structure guiding ground-level local maneuvers), and whether cross-view spatial reasoning transfers from static VQA-style diagnostics to sequential control where errors can accumulate over time. Addressing this requires benchmarks that couple synchronized heterogeneous views with both static diagnostics and closed-loop navigation evaluation.

\section{AirGroundBench Overview}

AirGroundBench evaluates the \textit{geometric consistency} and \textit{cross-view spatial intelligence} of multimodal large language models in heterogeneous air--ground embodied settings. Each sample contains synchronized observations captured at the same timestamp from two physically distinct platforms: an aerial drone (UAV) and a ground vehicle (UGV). The UAV view offers global structural context, while the UGV view provides fine-grained local details. This dual-view heterogeneity introduces systematic challenges---notably scale discrepancies, asymmetric occlusions, and reference-frame inconsistencies---making cross-view spatial reasoning substantially harder than conventional single-view perception.

\subsection{Environments and Data Scale}

We construct the benchmark in high-fidelity simulation environments spanning diverse outdoor scene types (e.g., urban districts, residential communities, coastlines, deserts, mountains, and industrial zones). In this paper, we evaluate models on 11 environments and collect 1,021 synchronized UAV--UGV image pairs. Based on these paired observations, AirGroundBench provides 62k multi-view VQA instances and 115 vision--language navigation (VLN) episodes. All VQA problems are 4-way single-choice questions.

\subsection{Tasks and Annotations}

AirGroundBench contains 10 tasks organized into four capability dimensions (Sec.~4): spatial perception, spatial alignment, spatial reasoning, and spatial decision-making. The first nine tasks are static multi-view VQA: each instance includes synchronized multimodal observations and a question about spatial relations, object attributes, or hypothetical scene conditions, requiring the model to select the correct option. The final task assesses decision-making via cross-view VLN, where the model iteratively selects actions from the current dual-view observations and a language instruction in a closed-loop environment.

To enable geometry-grounded evaluation, we provide structured spatial annotations for each observation pair, including cross-view object identities and 2D/3D bounding boxes in metric units. These annotations support quantitative supervision and diagnostic analyses of distance, direction, and relative pose.

\subsection{Evaluation Settings}

To isolate the contribution of heterogeneous views, we evaluate models under three controlled input configurations: \textit{UGV-only}, \textit{UAV-only}, and \textit{Dual-view (UAV+UGV)}. For the nine VQA tasks, we report answer accuracy; for VLN, we adopt standard metrics: Success Rate (SR), Success weighted by Path Length (SPL), and Navigation Error (NE). Full evaluation protocols are provided in Sec.~6 and the Appendix.

Overall, AirGroundBench is designed as a diagnostic framework that probes scale alignment, entity correspondence, reasoning under partial observability, and sequential embodied decision-making, and its task taxonomy (Sec.~4) provides a structured lens for analyzing these capabilities.

\section{Task Taxonomy}
\label{sec:taxonomy}

\subsection{Theoretical Motivation}

\begin{figure}[t]
  \centering
  \includegraphics[width=\linewidth]{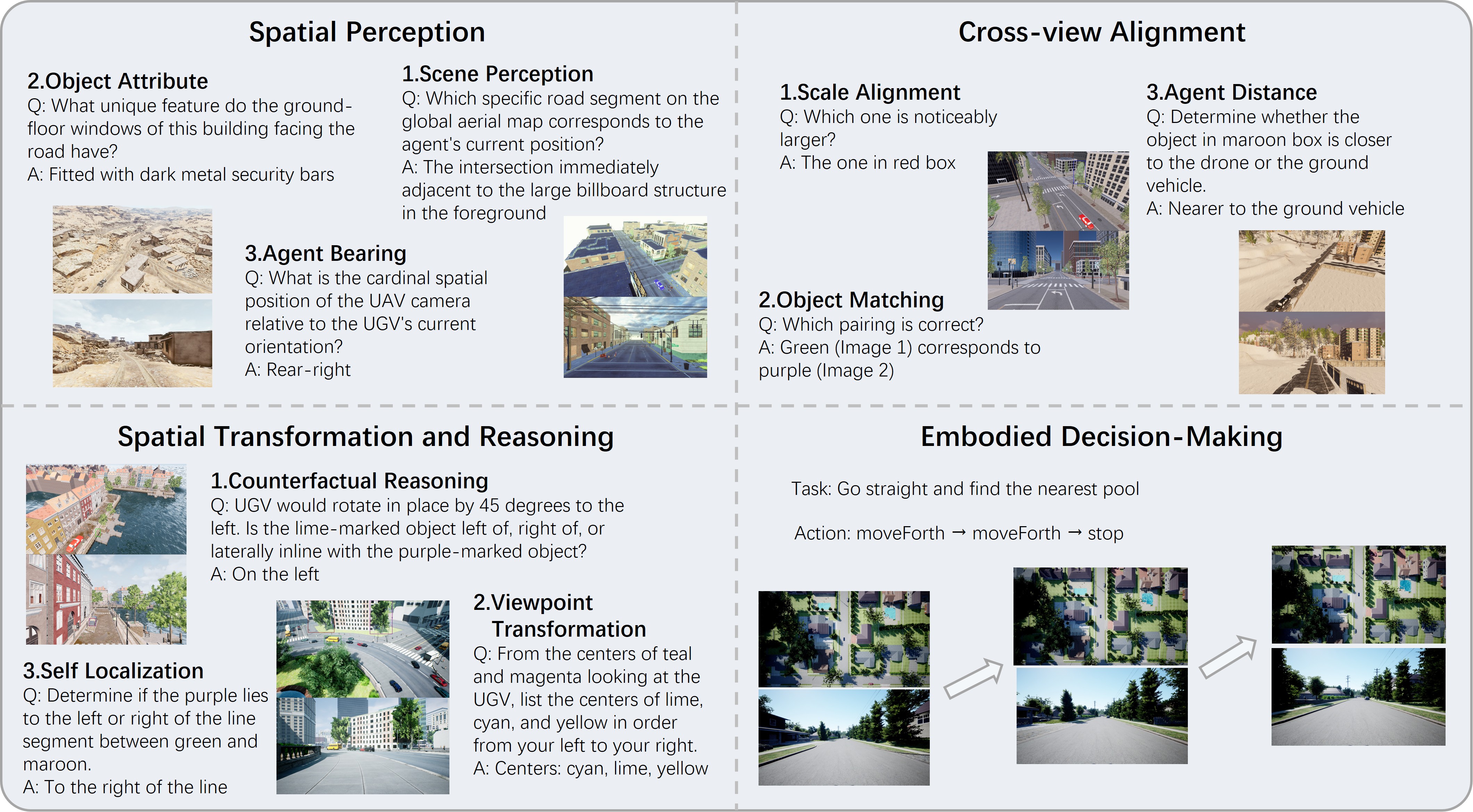}
  \caption{\textbf{Task taxonomy of AirGroundBench.} We organize 10 task types into four capability dimensions: spatial perception, cross-view alignment, spatial transformation and reasoning, and embodied decision-making. Each task is formulated to probe geometric consistency and cross-view spatial understanding under heterogeneous UAV--UGV observations.}
  \label{fig:taxonomy}
\end{figure}

Embodied spatial intelligence requires a composition of perceptual interpretation, cross-view correspondence, viewpoint transformation, and action-oriented spatial updating. Cognitive studies suggest that humans rely on mental rotation and internal transformations when comparing spatial configurations \cite{shepard1971mentalrotation}, incur measurable costs when switching reference frames \cite{hegarty2004dissociation}, and build cognitive maps to integrate local observations into coherent global representations \cite{tolman1948cognitivemaps,okeefe1978hippocampus}. From an embodied AI perspective, these processes can be viewed as maintaining geometric consistency under SE(3) transformations with partial observability, where small inconsistencies may accumulate into failures in sequential interaction. Motivated by these insights, AirGroundBench decomposes heterogeneous UAV--UGV spatial intelligence into four capability dimensions---spatial perception, cross-view alignment, spatial transformation and reasoning, and embodied decision-making---each instantiated by diagnostic tasks (Fig.~\ref{fig:taxonomy}).

\subsection{Spatial Perception}

Spatial perception assesses whether models can extract spatially meaningful cues from heterogeneous UAV--UGV observations without explicit geometric reconciliation.
\textbf{Scene Perception (P-Sce)} targets holistic environment understanding by integrating aerial structure with ground-level context.
\textbf{Object Attributes (P-Obj)} probes recognition of object-level properties under viewpoint/scale differences, typically combining aerial candidate localization with ground-level appearance cues.
\textbf{Agent Bearing (P-Bear)} evaluates relative bearing awareness of the UAV w.r.t.\ the UGV heading, requiring consistent interpretation of heading cues and cross-view configuration.

\subsection{Cross-view Alignment}

Cross-view alignment evaluates whether models can establish geometrically consistent correspondences between aerial and ground observations.
\textbf{Scale Alignment (A-Scale)} probes metric-scale reconciliation across views (e.g., size/distance ordering or directional relations aligned to the UGV heading), where correct inference requires consistent cross-view metric grounding.
\textbf{Object Matching (A-Match)} tests cross-view entity binding by identifying the corresponding instance across views under viewpoint asymmetry and occlusion.
\textbf{Agent Distance (A-Dist)} requires quantitative estimation of UAV--UGV relations (e.g., distance/height/relative direction), emphasizing metric-aligned dual-view geometry rather than qualitative relation naming.

\subsection{Spatial Transformation and Reasoning}

This dimension probes reasoning that goes beyond direct cross-view consistency, requiring spatial transformation, counterfactual inference, or multi-constraint geometric deduction.
\textbf{Counterfactual Reasoning (R-CF)} asks models to predict how spatial relations change under hypothetical state updates (e.g., agent motion or altitude changes).
\textbf{Viewpoint Transformation (R-View)} requires inferring relations from a virtual reference frame anchored at a specified object/location, demanding explicit viewpoint remapping.
\textbf{Self Localization (R-Loc)} formulates multi-constraint geometric localization (e.g., sidedness w.r.t.\ a landmark-defined line or inclusion within a triangle), testing robust spatial deduction under ambiguity.

\subsection{Embodied Decision-Making}

Vision--language navigation is a canonical and high-stakes embodiment of spatial intelligence, where perception, alignment, and transformation must be continuously maintained and translated into sequential actions.
\textbf{Vision-Language Navigation (VLN)} evaluates closed-loop control under synchronized UAV--UGV observations: at each step, the model selects an action from a discrete action space conditioned on the current dual-view observations and an instruction, and receives updated observations reflecting the new state. Successful navigation requires leveraging aerial global situational awareness together with ground-level local structure while preserving geometric consistency across time steps.

\section{Benchmark Construction}
\label{sec:construction}

\subsection{Simulation Platform and Data Capture}

\begin{figure}[t]
  \centering
  \begin{subfigure}[t]{0.32\linewidth}
    \centering
    \includegraphics[width=\linewidth]{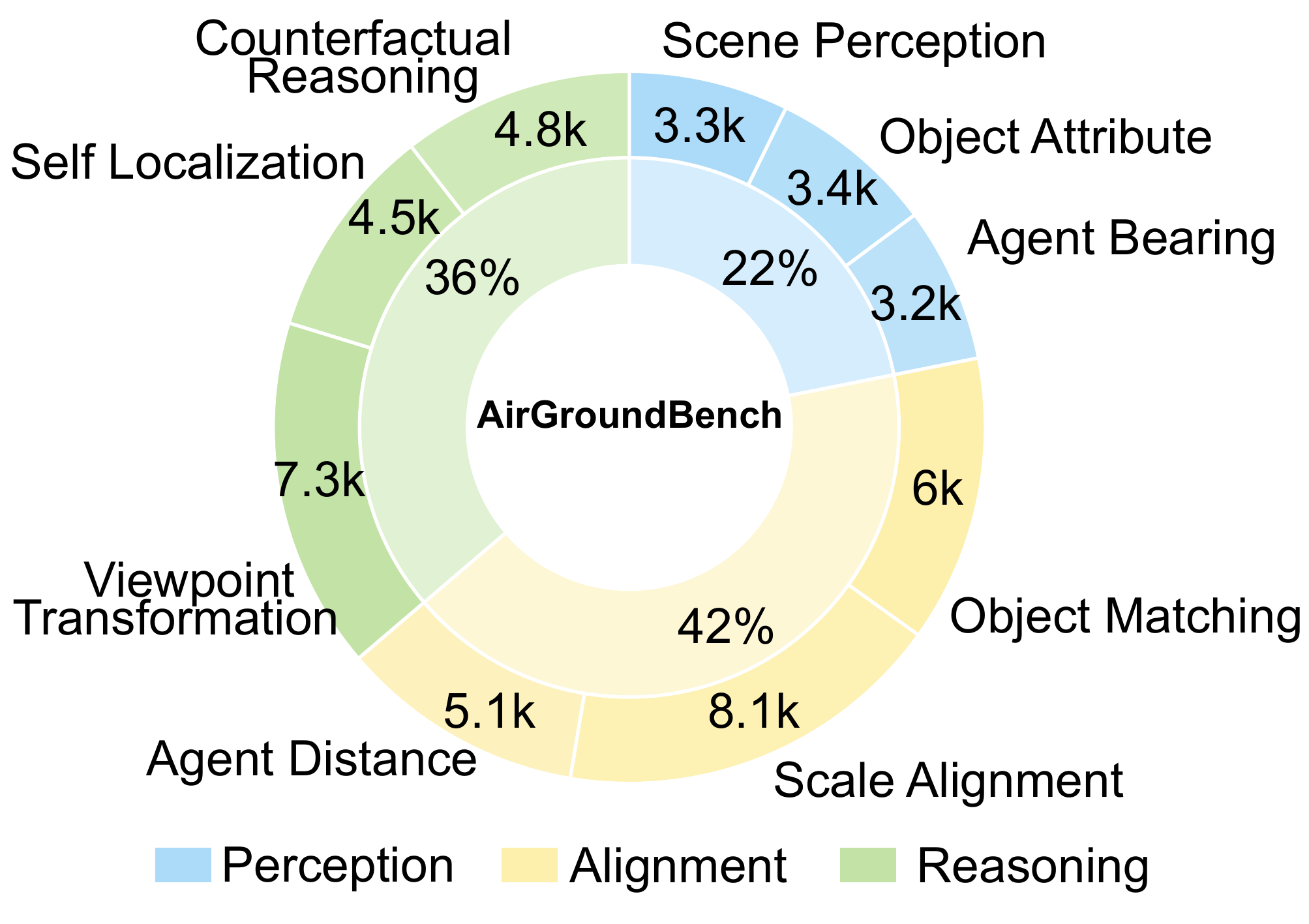}
    \caption{VQA task distribution.}
    \label{fig:dataset_stats_vqa}
  \end{subfigure}\hfill
  \begin{subfigure}[t]{0.32\linewidth}
    \centering
    \includegraphics[width=\linewidth]{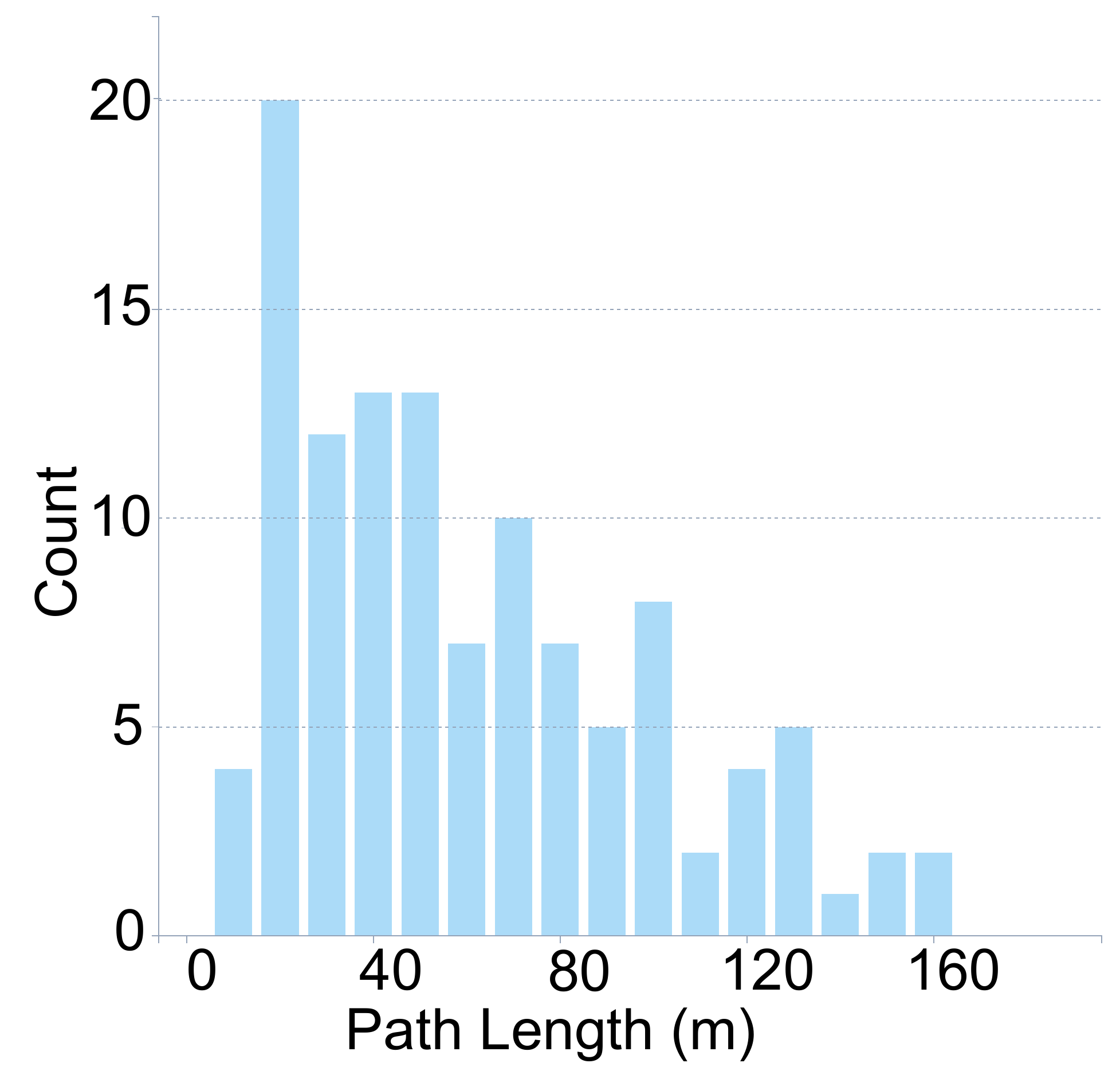}
    \caption{VLN path lengths.}
    \label{fig:dataset_stats_vln}
  \end{subfigure}\hfill
  \begin{subfigure}[t]{0.32\linewidth}
    \centering
    \includegraphics[width=\linewidth]{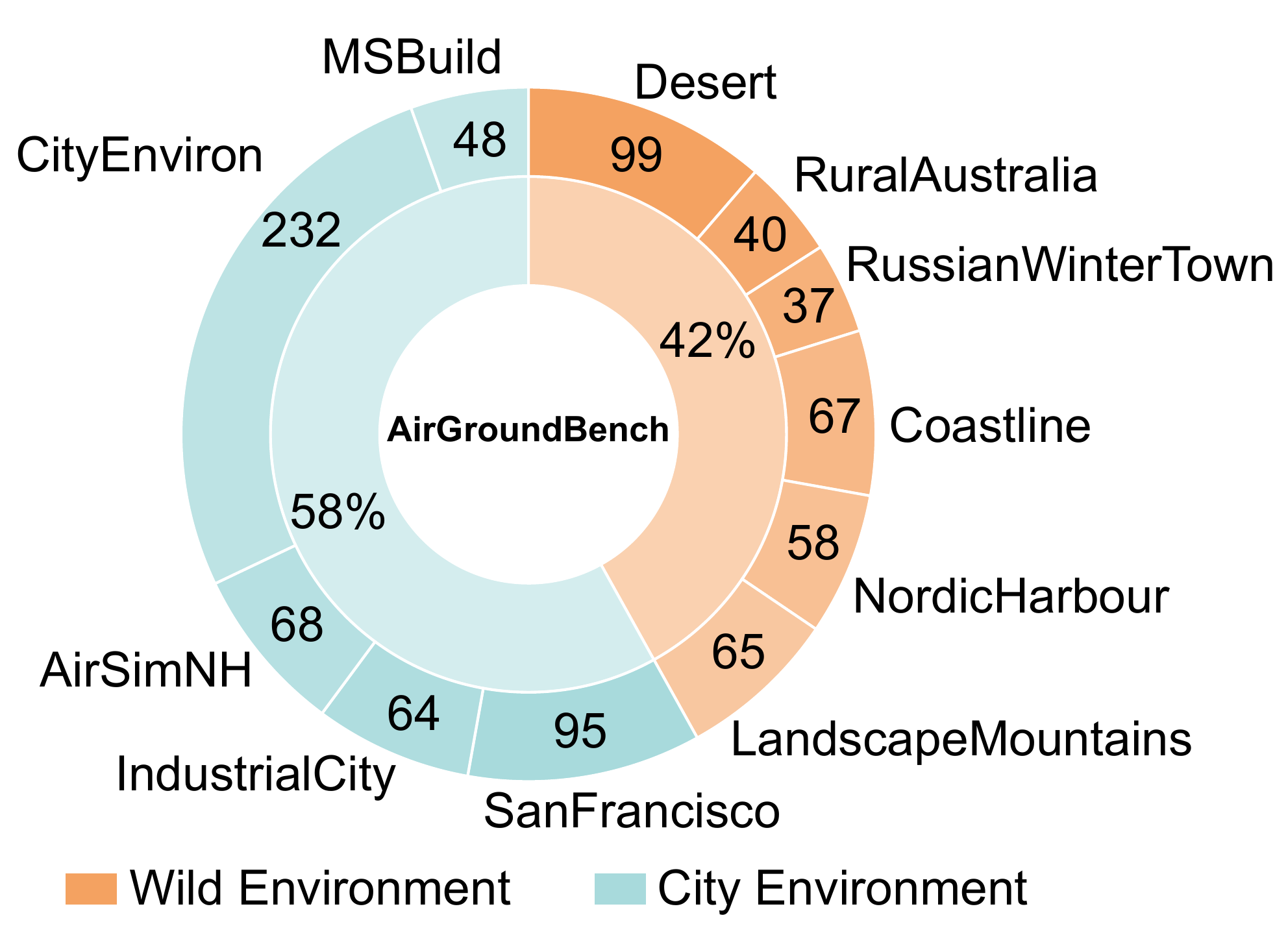}
    \caption{Environment composition.}
    \label{fig:dataset_stats_env}
  \end{subfigure}

  \caption{\textbf{AirGroundBench dataset overview.}
  (a) VQA question distribution across capability groups (Perception/Alignment/Reasoning) with per-task counts.
  (b) Distribution of shortest-path lengths for VLN episodes (meters).
  (c) Environment composition (City vs.\ Wild) and per-environment image-pair counts.}
  \label{fig:dataset_stats}
\end{figure}

AirGroundBench is built in a high-fidelity simulation stack based on Unreal Engine and AirSim, enabling synchronized control of a ground vehicle (UGV) and an aerial vehicle (UAV). For each timestamp, we record a temporally aligned dual-view observation pair, consisting of the UGV egocentric view and the UAV aerial view. Along with RGB frames, we also store auxiliary visual modalities (e.g., depth and segmentation) and the simulator states required for metric grounding and structured annotation. 
Figure~\ref{fig:dataset_stats} summarizes key statistics of the collected dataset, including the distribution of VQA questions across capability groups, the VLN path-length distribution, and the environment composition.

\subsection{Controlled Dual-View Sampling via UAV Viewpoint Presets}

To ensure systematic and reproducible cross-view geometry, UAV viewpoints are not sampled in global world coordinates. Instead, each UAV camera pose is specified by a preset defined in the UGV-local NED frame as a fixed relative transformation. Let $\mathbf{T}^{w}_{g}(t)\in SE(3)$ denote the UGV pose in world coordinates at time $t$, and $\mathbf{T}^{g}_{u}\in SE(3)$ a preset-defined UAV pose relative to the UGV. The UAV pose is obtained by composition:

\begin{equation}
\mathbf{T}^{w}_{u}(t)=\mathbf{T}^{w}_{g}(t)\,\mathbf{T}^{g}_{u}.
\end{equation}

This design directly controls the UAV--UGV baseline and viewpoint disparity, and makes cross-view scale calibration and identity binding comparable across environments. Presets are sampled with predefined weights to balance commonly useful cooperative views with harder wide-baseline cases; low-quality frames (e.g., artifacts or abnormal dynamics) are filtered during post-processing, and final distribution statistics are reported in the Appendix.

\subsection{Structured Spatial Annotation with Verifiable Geometry}

Each synchronized dual-view pair is annotated at collection time with structured spatial metadata, including object identifiers, 2D bounding boxes, and metric 3D bounding boxes. We additionally store the relative UAV--UGV pose metadata, enabling verifiable supervision signals for distance intervals, orientation relations, and cross-view consistency checks. This geometry-grounded annotation supports diagnostic evaluation beyond answer correctness.

\subsection{Question and Episode Generation with Quality Control}

We construct the benchmark via a semi-automatic pipeline: (i) template-driven generation grounded in structured annotations, (ii) model-assisted linguistic diversification, and (iii) human verification to ensure unambiguous and spatially grounded questions. All VQA problems are formulated as four-option single-choice questions. For navigation, we generate interactive VLN episodes under a closed-loop protocol with a discrete action space, and standard metrics (SR/SPL/NE) are computed from simulator trajectories. Additional quality checks include automated schema/consistency validation and manual audits; human evaluation details are deferred to the Appendix.

\section{Experiments}

\subsection{Experimental Setup}
\label{sec:exp_setup}

\subsubsection{Baselines}
\label{sec:baselines}

We evaluate a diverse set of multimodal large language models (MLLMs), including both proprietary systems and open-source baselines. 
Our evaluated models include: GPT-5.2~\cite{openai_gpt4v}, Gemini-3.1-Pro~\cite{gemini}, Claude-Sonnet-4.6~\cite{anthropic_claude}, Doubao-Seed-2.0~\cite{doubao}, ERNIE-5.0~\cite{ernie_bot}, GLM-4.6V~\cite{glm4v}, Hunyuan-Vision~\cite{hunyuan}, Kimi-K2.5~\cite{kimi_k2}, Qwen3.5 (27B and 397B variants) and Qwen3.5-Plus~\cite{qwen_vl}, as well as Qwen3-VL (8B and 235B variants)~\cite{qwen2vl}. 
We additionally report human performance as an upper bound and include a random policy as a sanity-check baseline.

All models are evaluated in a zero-shot setting without any fine-tuning on AirGroundBench data.

\subsubsection{Evaluation Metrics}
\label{sec:metrics}

For the nine VQA tasks, performance is measured using \textbf{classification accuracy}. Since each question is formulated as a four-option single-choice problem, accuracy provides a direct and interpretable measure of task success, consistent with standard VQA-style evaluation~\cite{antol2015vqa,hudson2019gqa}.

For cross-view vision-language navigation, we report three standard embodied navigation metrics: \textbf{Success Rate (SR)}, \textbf{Success weighted by Path Length (SPL)}, and \textbf{Navigation Error (NE)}~\cite{anderson2018vision,krantz2020beyond}. SR measures whether the agent reaches the goal region, SPL accounts for path efficiency relative to shortest paths, and NE measures the final distance to the goal.

\subsubsection{Evaluation Protocol}
\label{sec:protocol}

All models are evaluated under a unified inference protocol with standardized prompts and consistent multimodal inputs. 
For VQA tasks, each instance provides synchronized UAV and UGV observations at the same timestamp, together with the question and four answer options; the model outputs a single option as the final prediction.

For navigation, models interact with the environment in a closed-loop setting. At each step, the model receives the current synchronized UAV/UGV observations and the navigation instruction, and outputs an action from a discrete action space. Episode termination and success criteria follow the benchmark definition (details in the Appendix).

Unless otherwise specified, we report results under the dual-view setting (both UAV and UGV observations). We additionally evaluate restricted input settings as diagnostic controls in later analyses (Sec.~\ref{sec:view_ablation}) to quantify the contribution of multi-view information.

\begin{table*}[tb]
  \caption{\textbf{Main results on AirGroundBench.}
  We report the average accuracy over nine VQA tasks (Avg., \%) and per-task accuracy (\%) across four capability dimensions; for VLN, we report SR/SPL (\%) and NE (m; lower is better).
  The best result in each column is in \textbf{bold}, and the second-best is \underline{underlined}.}
  \label{tab:main_results}
  \centering
  \setlength{\tabcolsep}{3.1pt}
  \renewcommand{\arraystretch}{1.08}
  \scriptsize
  \resizebox{\linewidth}{!}{
  \begin{tabular}{@{}l|c|ccc|ccc|ccc|ccc@{}}
    \toprule
    \noalign{\vskip-3pt}
    \multicolumn{1}{c|}{} & \multicolumn{1}{c|}{} &
    \multicolumn{3}{c|}{\cellcolor{HeadPercep!70}\textbf{Perception}} &
    \multicolumn{3}{c|}{\cellcolor{HeadAlign!70}\textbf{Alignment}} &
    \multicolumn{3}{c|}{\cellcolor{HeadReason!70}\textbf{Reasoning}} &
    \multicolumn{3}{c}{\cellcolor{HeadVLN!70}\textbf{Decision}} \\
    Method & Avg. &
    \rotatebox{90}{P-Sce} & \rotatebox{90}{P-Obj} & \rotatebox{90}{P-Bear} &
    \rotatebox{90}{A-Scale} & \rotatebox{90}{A-Match} & \rotatebox{90}{A-Dist} &
    \rotatebox{90}{R-CF} & \rotatebox{90}{R-View} & \rotatebox{90}{R-Loc} &
    SR & SPL & $\downarrow$NE \\
    \noalign{\vskip-1.6pt}
    \midrule

    \noalign{\vskip-2.7pt}
    \rowcolor{BarBlue!70}
    \textit{\textbf{Baseline}} & & & & & & & & & & & & & \\
    Random & 25.24 &
    25.95 & 24.71 & 25.88 &
    24.73 & 25.57 & 25.34 &
    25.39 & 24.59 & 24.64 &
    3.48 & 2.86 & 58.81 \\
    Human & 80.93 &
    95.62 & 93.21 & 91.67 &
    76.67 & 74.37 & 77.98 &
    73.33 & 77.78 & 78.13 &
    83.87 & 68.37 & 11.50 \\
    \noalign{\vskip-1.6pt}
    \midrule

    \noalign{\vskip-2.7pt}
    \rowcolor{BarBlue!70}
    \textit{\textbf{Proprietary Models}} & & & & & & & & & & & & & \\

    Doubao-Seed-2.0-Pro & \textbf{57.14} &
    93.10 & \underline{68.97} & 37.93 &
    \underline{56.12} & \textbf{56.83} & 65.13 &
    44.62 & \textbf{44.00} & \textbf{66.23} &
    \textbf{38.71} & \textbf{31.45} & \textbf{23.10} \\

    Qwen3.5-Plus & \underline{55.42} &
    86.21 & 65.52 & \textbf{48.28} &
    \textbf{59.57} & 43.84 & \underline{68.52} &
    \textbf{48.57} & 41.18 & \underline{56.76} &
    19.02 & 16.34 & 39.98 \\

    Kimi-K2.5 & 49.09 &
    89.66 & 58.62 & \underline{38.63} &
    48.23 & 40.00 & 64.16 &
    40.30 & 38.24 & 55.41 &
    25.85 & 19.75 & 38.94 \\

    Gemini-3.1-Pro & 46.79 &
    86.21 & 44.83 & 20.69 &
    49.68 & 50.34 & \textbf{72.28} &
    43.42 & 22.40 & 41.11 &
    26.32 & 22.87 & 42.17 \\

    Claude-Sonnet-4.6 & 46.16 &
    \textbf{96.86} & 65.52 & 17.24 &
    48.92 & 28.06 & 56.39 &
    44.35 & \underline{39.08} & 56.36 &
    26.69 & 20.80 & \underline{29.09} \\

    GPT-5.2 & 45.89 &
    89.66 & \textbf{79.31} & 20.69 &
    44.60 & 47.48 & 50.15 &
    \underline{44.93} & 29.41 & 43.24 &
    25.22 & 19.99 & 47.34 \\

    GLM-4.6V & 45.76 &
    \underline{96.55} & 55.17 & 27.59 &
    44.60 & 38.03 & 62.45 &
    38.46 & 26.73 & 56.76 &
    19.25 & 15.58 & 36.20 \\

    ERNIE-5.0 & 44.67 &
    86.21 & 55.17 & 37.93 &
    38.67 & \underline{50.35} & 59.62 &
    35.82 & 24.30 & 44.74 &
    \underline{32.26} & \underline{25.32} & 43.34 \\

    Hunyuan-Vision & 34.42 &
    89.66 & 31.03 & 24.14 &
    32.62 & 35.25 & 46.48 &
    29.23 & 26.13 & 20.27 &
    6.45 & 6.14 & 45.57 \\
    \noalign{\vskip-1.6pt}
    \midrule

    \noalign{\vskip-2.7pt}
    \rowcolor{BarBlue!70}
    \textit{\textbf{Open-source Models}} & & & & & & & & & & & & & \\

    Qwen3.5-397B & \textbf{55.11} &
    \underline{93.10} & \underline{68.63} & \textbf{45.26} &
    \textbf{53.24} & \textbf{46.04} & \textbf{71.35} &
    \textbf{45.71} & \textbf{48.35} & \textbf{57.12} &
    19.09 & 15.79 & \textbf{38.51} \\

    Qwen3.5-27B & \underline{52.70} &
    89.66 & \textbf{68.79} & \underline{44.83} &
    \underline{49.64} & \underline{43.17} & \underline{69.35} &
    \underline{43.25} & \underline{43.69} & \underline{56.32} &
    \underline{28.78} & \underline{22.68} & 45.78 \\

    Qwen3-VL-235B & 44.32 &
    92.17 & 55.54 & 41.38 &
    45.32 & 38.13 & 64.58 &
    35.38 & 24.15 & 40.54 &
    \textbf{29.39} & \textbf{23.80} & \underline{39.07} \\

    Qwen3-VL-8B & 42.19 &
    \textbf{96.63} & 55.17 & 27.59 &
    35.25 & 42.45 & 63.08 &
    35.19 & 24.74 & 36.49 &
    16.13 & 13.34 & 40.30 \\
    \noalign{\vskip-1.6pt}
    \bottomrule
  \end{tabular}}
\end{table*}

\subsection{Main Results}
\label{sec:main_results}

Table~\ref{tab:main_results} summarizes the main results under the dual-view setting across 10 tasks grouped into four capability dimensions. We highlight three takeaways that are most relevant to the benchmark’s motivation.

\textbf{(1) Large headroom remains: dual-view spatial intelligence is far from solved.}
Even the strongest MLLMs reach only mid-50s average accuracy on the nine VQA tasks, leaving a substantial gap to the human reference ($\sim$80\% on VQA). This indicates that heterogeneous UAV--UGV multi-view understanding is not yet reliably captured by current models, despite their strong general vision-language competence.

\textbf{(2) The dominant bottleneck is geometric consistency across views.}
Performance is comparatively higher on Perception tasks, where complementary aerial context and ground-level details support recognition. In contrast, results drop most consistently on Cross-view Alignment and remain unstable on Spatial Transformation and Reasoning. In particular, tasks that require metric-scale consistency, cross-view identity binding, or reference-frame remapping exhibit the most pronounced degradations, suggesting that models often fail to form a coherent, view-consistent geometric representation.

\textbf{(3) Closed-loop navigation exposes error accumulation.}
The VLN setting is substantially harder than single-step VQA: success rates remain far below the human reference, and trajectory efficiency is limited even in successful episodes. This gap is consistent with an accumulation effect: small cross-view inconsistencies that might be tolerated in one-shot predictions become consequential when iterated over time. Overall, VLN serves as a stringent end-to-end test that reveals whether a model’s spatial understanding is action-consistent rather than merely descriptive.

\begin{table*}[tb]
\caption{\textbf{Scene generalization across environment types on AirGroundBench.}
Environments are grouped into \emph{City} and \emph{Wild}, and we report the average accuracy over the nine VQA tasks (Avg., \%) for each evaluated environment.
The best result in each column is in \textbf{bold}, and the second-best is \underline{underlined}.}
\label{tab:scene_generalization}
\centering

\setlength{\tabcolsep}{2.6pt}
\renewcommand{\arraystretch}{1.12}
\small

\resizebox{\linewidth}{!}{
\begin{tabular}{l|c|ccccc|cccccc}
\toprule

\noalign{\vskip-3pt}
& & \multicolumn{5}{c|}{\cellcolor{CitySim!70}\textbf{City Environment}}
& \multicolumn{6}{c}{\cellcolor{WildSim!70}\textbf{Wild Environment}} \\

Method & Avg.
& CityEnv & AirSim & IndCity & SanFran & MSBuild
& MtLand & Nordic & Coast & RusWin & RuralAus & Desert \\

\noalign{\vskip-1.6pt}
\midrule

\noalign{\vskip-2.7pt}
Doubao-Seed-2.0-Pro
& \textbf{57.31}
& 49.76 & 63.05 & \textbf{56.80} & \textbf{57.98} & \textbf{47.15}
& \underline{50.21} & \textbf{62.24} & \textbf{42.30} & 45.58 & \underline{70.58} & \textbf{58.96} \\

Qwen3.5-Plus
& \underline{55.52}
& 45.67 & 57.93 & \underline{56.55} & \underline{57.56} & 41.34
& \textbf{57.76} & 59.72 & 42.04 & \underline{58.83} & 49.77 & 54.87 \\

Qwen3.5-397B
& 55.16
& 45.18 & \underline{63.35} & 55.70 & 56.94 & 40.94
& 45.93 & \underline{60.18} & 38.40 & 48.67 & 62.33 & 56.11 \\

Qwen3.5-27B
& 52.47
& 50.13 & 52.64 & 51.77 & 50.84 & 35.30
& 42.46 & 58.59 & \underline{42.13} & 39.39 & \textbf{70.98} & \underline{56.17} \\

Kimi-K2.5
& 48.91
& \textbf{63.60} & 47.67 & 45.87 & 45.36 & 35.40
& 46.00 & 44.74 & 34.39 & 54.44 & 54.00 & 55.67 \\

Gemini-3.1-Pro
& 46.95
& 45.69 & 58.17 & 44.11 & 49.15 & 41.02
& 46.27 & 48.07 & 34.57 & 58.63 & 48.07 & 45.52 \\

Claude-Sonnet-4.6
& 46.22
& 45.39 & 57.77 & 45.37 & 44.81 & \underline{47.13}
& 38.59 & 53.12 & 30.58 & 51.37 & 62.65 & 43.31 \\

GPT-5.2
& 46.11
& 40.76 & 47.43 & 49.03 & 46.30 & 41.30
& 34.63 & 52.42 & 31.02 & \textbf{60.79} & 58.23 & 41.85 \\

GLM-4.6V
& 45.81
& 50.25 & \textbf{68.46} & 43.29 & 37.86 & 41.13
& 34.76 & 51.96 & 42.03 & 44.11 & 53.94 & 48.66 \\

ERNIE-5.0
& 44.73
& 54.72 & 52.39 & 35.76 & 38.39 & 40.92
& 42.08 & 55.29 & 42.03 & 56.13 & 62.53 & 44.60 \\

Qwen3-VL-235B
& 44.59
& \underline{54.84} & 42.26 & 39.53 & 45.07 & 35.05
& 42.15 & 44.67 & 34.87 & 57.85 & 41.75 & 45.52 \\

Qwen3-VL-8B
& 42.02
& 45.71 & 36.99 & 37.90 & 41.44 & 35.14
& 30.51 & 44.24 & 34.66 & 57.69 & 50.26 & 43.69 \\

Hunyuan-Vision
& 34.43
& 27.10 & 47.27 & 34.89 & 35.12 & 23.53
& 22.98 & 39.05 & 31.06 & 45.26 & 45.98 & 31.24 \\

\noalign{\vskip-1.6pt}
\bottomrule
\end{tabular}}
\end{table*}

\subsection{Scene Generalization}
\label{sec:scene_generalization}

To evaluate how spatial understanding generalizes across environment structures, we group the evaluated environments into City and Wild categories and report the average accuracy over the nine VQA tasks in Table~\ref{tab:scene_generalization}. The results shows that performance varies not only across models but also across environment structures.

Overall, models exhibit noticeable variation across scenes. While the average accuracy between city and wild environments is comparable, the variance among wild scenes is significantly larger. In particular, environments characterized by large open spaces tend to produce lower performance across most models.

One likely reason is that such environments provide fewer stable geometric anchors. Without strong structural cues such as buildings, road layouts, or clear spatial boundaries, it becomes more difficult for models to establish consistent correspondences between aerial and ground observations.

In contrast, environments with clearer structural layouts generally yield higher and more stable performance. These results suggest that current MLLMs rely heavily on structural cues to anchor spatial representations, highlighting the importance of geometric grounding for robust multi-view spatial understanding.

\begin{table}[t]
\caption{\textbf{Viewpoint ablation on AirGroundBench.}
Models are evaluated under three input settings: UAV-only, UGV-only, and dual-view (UAV+UGV).
We report VQA performance in percentage (\%) for Avg./P/A/R and VLN success rate (SR; reported in the 0--1 range).}
\label{tab:view_ablation}
\centering
\setlength{\tabcolsep}{4pt}
\renewcommand{\arraystretch}{1.08}
\scriptsize
\begin{tabular}{l c c c c c c}
\toprule
\textbf{Model} & \textbf{View} & \textbf{Avg.} & \textbf{P} & \textbf{A} & \textbf{R} & \textbf{SR} \\
\midrule

\multirow{3}{*}{Doubao-Seed-2.0-Pro}
& UAV  & 38.68 & 55.56 & 36.90 & 34.29 & 0.25 \\
& UGV  & 34.91 & 64.44 & 33.33 & 24.76 & 0.26 \\
& BOTH & \textbf{57.14} & 66.67 & \textbf{58.73} & \textbf{51.24} & \textbf{0.39} \\
\midrule

\multirow{3}{*}{Qwen3.5-Plus}
& UAV  & 40.94 & 61.70 & 37.50 & 37.14 & 0.08 \\
& UGV  & 37.50 & 57.45 & 32.14 & 37.14 & 0.11 \\
& BOTH & \textbf{55.42} & 66.67 & \textbf{55.81} & \textbf{50.81} & \textbf{0.19} \\
\midrule

\multirow{3}{*}{Qwen3.5-397B}
& UAV  & 40.25 & 57.78 & 33.33 & 43.81 & 0.07 \\
& UGV  & 35.42 & 63.04 & 30.95 & 30.48 & 0.11 \\
& BOTH & \textbf{55.11} & 68.97 & \textbf{55.29} & \textbf{49.79} & \textbf{0.19} \\
\midrule

\multirow{3}{*}{Qwen3.5-27B}
& UAV  & 43.81 & 53.19 & 40.24 & 45.19 & 0.17 \\
& UGV  & 38.87 & 55.10 & 38.55 & 31.73 & 0.19 \\
& BOTH & \textbf{52.70} & 67.82 & \textbf{52.38} & \textbf{47.70} & \textbf{0.29} \\
\midrule

\multirow{3}{*}{Kimi-K2.5}
& UAV  & 44.66 & 51.16 & 43.37 & 44.00 & 0.16 \\
& UGV  & 35.50 & 53.49 & 27.61 & 40.59 & 0.12 \\
& BOTH & \textbf{49.09} & 62.07 & \textbf{49.34} & \textbf{44.03} & \textbf{0.26} \\

\bottomrule
\end{tabular}
\end{table}

\subsection{Viewpoint Ablation}
\label{sec:view_ablation}

To examine the contribution of heterogeneous viewpoints, we evaluate models under three input configurations: \textit{UAV-only}, \textit{UGV-only}, and \textit{dual-view} (UAV+UGV). Results are summarized in Table~\ref{tab:view_ablation}.

Across all evaluated models, dual-view input consistently yields the best performance. For example, Doubao-Seed-2.0-Pro improves from 38.68\% (UAV-only) and 34.91\% (UGV-only) to 57.14\% under dual-view input. Similar trends are observed for all other models. This result confirms that aerial and ground observations provide complementary spatial information that benefits multi-view reasoning.

Comparing single-view settings reveals distinct advantages of each viewpoint. UAV observations typically provide stronger scene-level cues due to their global coverage, while UGV observations often yield better object-level recognition owing to their proximity to scene entities. This complementary information explains the consistent gains observed when both viewpoints are provided.

The effect is particularly pronounced in embodied navigation. For instance, the success rate of Doubao-Seed-2.0-Pro increases from 0.25 (UAV-only) and 0.26 (UGV-only) to 0.39 under dual-view input. This suggests that multi-view spatial integration becomes increasingly important when spatial reasoning must be maintained over sequential decision steps.

\subsection{Diagnostic Analysis}

To better understand the limitations of current MLLMs on heterogeneous multi-view spatial reasoning, we analyze performance across the four capability dimensions defined in our taxonomy.

A clear capability hierarchy emerges across evaluated models: spatial perception generally achieves the highest accuracy, followed by cross-view alignment, spatial reasoning, and finally embodied decision-making. This progressive degradation suggests that while current models can reliably extract semantic information from individual observations, maintaining geometric consistency across heterogeneous viewpoints remains substantially more difficult.

The gap becomes more pronounced in tasks that require explicit geometric transformation. Alignment tasks such as scale calibration and cross-view matching already introduce noticeable performance drops, indicating that models struggle to construct a shared metric representation across aerial and ground perspectives. Reasoning tasks further amplify this difficulty, as they require coordinate-frame transformation or ambiguity resolution under partial observability.

The largest discrepancy appears in embodied navigation. While human participants achieve approximately 90\% success rate, the best-performing model remains below 40\%. This gap suggests that small spatial inconsistencies accumulated during perception and alignment stages can propagate through sequential decision-making, ultimately leading to navigation failure.

Overall, these observations indicate that the primary bottleneck for current MLLMs lies not in recognizing visual content, but in maintaining geometrically consistent spatial representations across heterogeneous viewpoints and over time.

\section{Conclusion}

We introduced AirGroundBench, a benchmark for evaluating heterogeneous dual-view spatial intelligence in embodied settings. Built from synchronized UAV–UGV observations across diverse simulated environments, the benchmark systematically evaluates spatial perception, cross-view alignment, spatial reasoning, and embodied decision-making through nine VQA tasks and a vision-language navigation task. Extensive experiments on a wide range of multimodal large language models reveal a clear capability hierarchy: while models perform relatively well on perceptual understanding, their performance degrades substantially when cross-view geometric alignment and coordinate transformation are required, with the largest gap appearing in embodied navigation. These findings suggest that maintaining geometrically consistent spatial representations across heterogeneous viewpoints remains a fundamental challenge for current MLLMs. We hope AirGroundBench will serve as a diagnostic benchmark for future research on multi-view spatial reasoning and embodied intelligence.

%
%
\clearpage
\bibliographystyle{splncs04}
\bibliography{main, zot, relatedwork}

\end{document}